# A compact aVLSI conductance-based silicon neuron

Runchun Wang, Chetan Singh Thakur, Tara Julia Hamilton, Jonathan Tapson, André van Schaik
The MARCS Institute, University of Western Sydney, Sydney, NSW, Australia
mark.wang@uws.edu.au

*Abstract*—We present an analogue Very Large Scale Integration (aVLSI) implementation that uses first-order low-pass filters to implement a conductance-based silicon neuron for high-speed neuromorphic systems. The aVLSI neuron consists of a soma (cell body) and a single synapse, which is capable of linearly summing both the excitatory and inhibitory post-synaptic potentials (EPSP and IPSP) generated by the spikes arriving from different sources. Rather than biasing the silicon neuron with different parameters for different spiking patterns, as is typically done, we provide digital control signals, generated by an FPGA, to the silicon neuron to obtain different spiking behaviours. The proposed neuron is only ~26.5 µm$^2$ in the IBM 130nm process and thus can be integrated at very high density. Circuit simulations show that this neuron can emulate different spiking behaviours observed in biological neurons.

## I. Background

Conductance-based neuron models [1] are easier to implement in hardware than the full Hodgkin-Huxley model [2], while still being rich in behaviours observed in biological neurons. The log-domain LPF neuron [3] and the Diff-Pair Integrator neuron [3] represent good examples of conductance-based silicon neurons. These neurons operate in the log-domain and use only a few transistors and are capable of reproducing a variety of spiking behaviours, such as regular spiking, spiking frequency adaptation and bursting.

In our previous work [4], we have presented an aVLSI conductance-based silicon neuron that occupies ~170 µm$^2$ and is rich in dynamic behaviours that are similar to its biological counterpart. Here we present an improved aVLSI conductance-based silicon neuron, which occupies only ~26.5 µm$^2$ without sacrificing any performance. A test chip containing 0.5 million of these neurons has been fabricated using the IBM 130nm technology. In this paper, we present the structure of the proposed silicon neuron along with circuit simulation results. The outline for this paper is as follows. The CMOS implementation is presented in Section II. In Section III we present the simulation results and we conclude in Section IV.

## II. CMOS Implementation

### A. Motivation

Most of the implementations of existing conductance-based silicon neurons have to be tuned with specified parameters, some of which are very sensitive to the error, to model the dynamics of the neuron's membrane voltage for different types of biological neurons. In this way, these silicon neurons will reproduce different firing patterns. This approach is time consuming and inefficient; it is only an option for a system with a small number of neurons, e.g., less than one hundred neurons. Due to process variation and device mismatch, especially for deep sub-micron technologies, it is impractical to tune all the neurons in large-scale spiking neural networks with the individual parameters needed to obtain the required behaviour. An ideal silicon neuron should be able to work with one and only one set of fixed but configurable parameters for instantiations on the Integrated Circuit (IC).

Programmable devices such as FPGAs may be employed for the generation of control signals that could be used by the neurons for achieving different firing patterns. Moreover, this scheme significantly reduces the size of the silicon neuron since it removes the on-chip feedback generating circuits, which usually need large capacitors to operate on biological time scales with typical time constants in the order of tens to hundreds of milliseconds. Our proposed scheme does, however, require a controller implemented on an FPGA. This does not increase the cost of the system significantly as, in state-of-the-art neuromorphic systems, an FPGA is needed anyway to carry out other tasks such as routing the spikes between analogue computation modules and other miscellaneous tasks.

### B. Design choices

In deep sub-micron technologies, generally, there are two types of devices: thin- and thick-oxide transistors. The latter have larger feature sizes, can use a higher power supply, have a lower switching speed, and are less sensitive to process variations than the former and are therefore usually used in aVLSI circuits. We used these in our previous implementation [4] and we achieved a density of 5882 neurons/mm$^2$. To further increase the density, we chose to use the thin-oxide transistors in the work reported here. The smaller feature size leads to a higher density, but these transistors are unsuitable to subthreshold circuits [5], which are most sensitive to process variation whether they are used in voltage or current mode [6]. Therefore, we use voltage mode low-pass filters in strong-inversion [5] to implement the neuron. Hence the proposed implementation is more robust than the log-domain implementations. Furthermore, this implementation uses fewer transistors than our previous work [4], which increases the density of neurons we can achieve. By leveraging the thin-oxide low-power devices of the IBM 130nm technology, a ~100 times leakage current reduction was achieved compared to the design with regular thin-oxide devices.

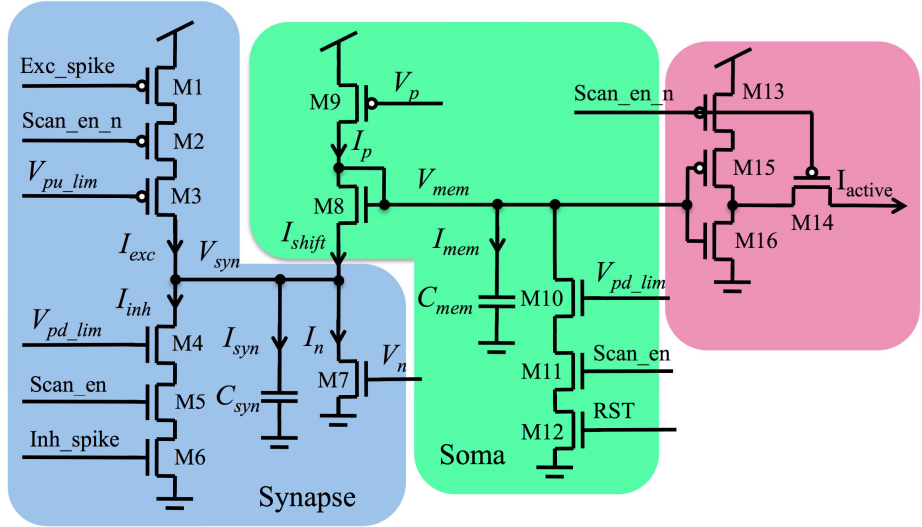

**Figure 1.** Schematic of the conductance-based neuron.

The schematic of the proposed silicon neuron is shown in Figure 1. The circuit comprises a conductance-based synapse (blue, M1-M7 and a MOS capacitor $C_{syn}$), a conductance-based soma (green, M8-M12 and a MOS capacitor $C_{mem}$) and an AER interface circuit (red, M13-M16). The proposed silicon neuron is placed in a neuron array and therefore requires selecting signals (Scan_en and Scan_en_n). For simplicity, we will assume the neuron in Figure 1 is always being selected and the transistors controlled by these signals are all ON in the following descriptions.

*C. Synapse*

The conductance-based synapse is implemented using a truly linear low-pass filter circuit, which can be shared to linearly sum the post-synaptic potentials (PSP) generated by the spikes arriving from different sources. Hence, only one such synaptic circuit is needed per neuron, which can save a significant amount of area on a chip. Synaptic plasticity, e.g. Spike Timing Dependent Plasticity (STDP) [7], could be carried out as in [8], which modifies the weights of the incoming spikes externally. Instead of using a separate low-pass filter for excitatory and inhibitory inputs, we proposed a conductance-based synapse that consists of one and only one synapse capacitor $C_{syn}$, a charge circuit (M1-M3), a discharge circuit (M4-M6) and M7 that models the neuron's conductance leak. This makes the circuit even more compact, but the drawback is that the time constant of the EPSP and IPSP will have to be identical.

The synapse behaves as an RC-filter. The voltage at node $V_{syn}$ will be increased/decreased by the current $I_{exc}$ and $I_{inh}$ that are generated by incoming excitatory and inhibitory spikes. The durations of excitatory and inhibitory spikes are modulated according to their weights with a gain set by the gate bias voltages $V_{pu\_lim}$ and $V_{pd\_lim}$ respectively. The current $I_n$ (~2 pA, depending on $V_{syn}$ due to channel length modulation [5]) is a leakage current set by the gate bias voltage $V_n$ and $I_{shift}$ is a current from the soma circuit, which will be presented in the next section. Kirchhoff's current law on the $V_{syn}$ node yields:

$$C_{syn}\frac{d}{dt}V_{syn} = I_{syn} = I_{exc} - I_{inh} - I_n - I_{shift} \quad (1)$$

After being increased or decreased by incoming spikes, $V_{syn}$ will be pulled up/down to a steady state such that $I_n = I_{shift}$. This is a self-balanced procedure since the value of $I_n$ will increase when $V_{syn}$ increases and vice versa.

*D. Soma*

The soma is implemented using another low-pass filter to model the neuron's leak conductance and the neuron's membrane capacitance (represented by $C_{mem}$). To feed the post-synaptic potential $V_{syn}$ to the soma circuit, we connect these two nodes via a level shifter circuit (a diode-connected transistor M8) such that the membrane voltage follows the first-order dynamics of the synapse. The current $I_p$ (~2 pA, depending on $V_{mem}$ due to channel length modulation) is a leakage current set by the gate bias voltage $V_p$. Ignoring the effect of the current from M10, Kirchhoff's current law on the $V_{mem}$ node yields:

$$C_{mem}\frac{d}{dt}V_{mem} = I_{mem} = I_p - I_{shift} \quad (2)$$

When $V_{syn}$ is charged by incoming excitatory spikes, M8 will be OFF ($I_{shift} = 0$) and $I_p$ will begin to pull $V_{mem}$ up until $I_p = I_{shift}$ again (M8 will be ON again since meanwhile $V_{syn}$ is being pulled down by $I_n$). After that, $V_{mem}$ will be pulled down by $I_{shift} - I_p$, while $V_{syn}$ is being pulled down by $I_n - I_{shift}$, to the steady state such that $I_p = I_{shift} = I_n$. When $V_{syn}$ is discharged by incoming inhibitory spikes, $V_{mem}$ will be pulled down by $I_{shift}$ very quickly (since $V_{syn}$ will be much smaller than $V_{mem}$ and hence $I_{shift} \gg I_p$) until $V_{mem}$ is low enough such that $I_{shift}$ is equal to $I_p$ (M8 is almost OFF). After that, $V_{mem}$ will be pulled up by $I_p$ - $I_{shift}$, meanwhile $V_{syn}$ is being pulled up by $I_{shift}$ - $I_n$, to the steady state such that $I_p = I_{shift} = I_n$.

The AER interface below will create spikes when $V_{mem}$ crosses a threshold. To model resetting of the membrane potential after a spike, the circuit of M10-M12 has been added. $V_{mem}$ will be pulled down by an incoming RST signal, the duration of which is modulated according to its strength with a gain set by the gate bias voltage $V_{pd\_lim}$.

### E. AER interface

In our previous work [9]–[11], we have presented a synchronous AER protocol optimised for a time-multiplexing architecture, which uses a collision-free serial scanning scheme with one active signal and an address. This synchronous scheme eliminates the overhead of an arbiter, which consumes a significant amount of silicon area in the standard AER protocol [12]. Hence we will apply this synchronous AER protocol to the proposed neuron with a scanning period of 1 ms, given that a millisecond resolution is generally acceptable for neural simulations.

The AER interface works as follows: when a neuron is selected, the voltage at node $V_{mem}$ in the soma circuit of that neuron will pull the output signal of this neuron, $I_{active}$, either up to $V_{dd}$ (inactive) or down to ground (active) via an inverter (M15-M16, for increasing driving capability) and switches (M13-M14). M13 is added to limit the power consumption when the neuron is not selected. When this neuron is not selected (M14 is OFF), the output node of this neuron will have high impedance, which releases the AER bus to be driven by other neurons in the neuron array. Once the off-chip FPGA reads the output signal as active, it will send feedback signals, e.g., the RST signal, back to the neuron depending on what firing pattern this neuron is configured for.

### F. Layout

All transistors are 280 nm wide and 280 nm long (M1-2, M5-6, M11-12 are 120 nm long). The inverter and switches (M13-M16) use transistors that are 720 nm wide and 120 nm long. The MOS capacitor values are: $C_{syn}$ = 1.4 μm x 1.42 μm (~22 fF), $C_{mem}$ = 1.4 μm x 1.22 μm (~18 fF). For the maximum utilisation of silicon area, a neuron should share as many resources with its neighbouring ones as possible. Based on this principle, all the pMOS and nMOS transistors of one neuron are located in the left and right side respectively. In this way the neurons can share their bulk connections (especially the ones of the MOS capacitors) with each other.

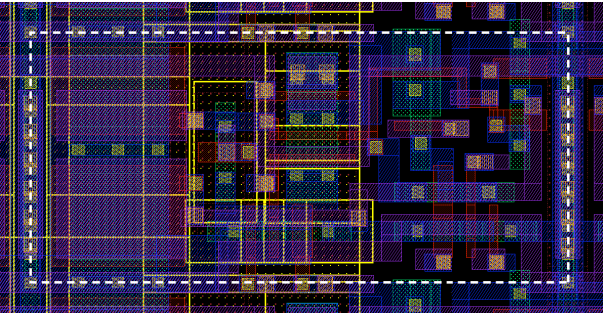

**Figure 2.** Layout of one conductance-based neuron.

The effective size of one neuron (see the white dashed rectangle in Figure 2) is ~26.5 μm$^2$ and the density that we achieved is 37,735 neurons/mm$^2$.

### III. SIMULATION RESULTS

We have run post-layout simulations for the proposed silicon neuron. In the simulations, the leak current $I_p$ and $I_n$ are set by biasing the gate voltage $V_p$ = 1.172 V and $V_n$ = 166.2 mV respectively. $V_{dd}$ was 1.2 V. The amplitude of the inhibitory and excitatory spikes are both set to 550 nA by biasing the gate voltages $V_{pd\_lim}$ = 645 mV and $V_{pu\_lim}$ = 628 mV respectively. These settings will remain the same for all spiking patterns shown below.

Figure 3 shows the simulation results. In Figure 3a the circuit behaves as a simple leaky-integrate-and-fire neuron without adaptation. Note that the glitches, which appear periodically (every 1 ms) on $V_{syn}$ (the green curve) and $V_{mem}$ (the magenta curve), are caused by digital noise introduced through capacitive coupling by the serial scanning signals (scan_en and scan_en_n, the durations of which were set to 32 ns). The EPSPs generated by four consecutive excitatory input spikes (the red curve, each with a duration of 24 ns and the inter-spike interval is set to 1 ms) are just enough to cause the neuron to spike (the blue curve, an inverted version of $I_{active}$). This spike will be scanned by an FPGA, and in simulation, we used a HDL model for this. The FPGA will then send the RST signal, the duration of which was set to 24 ns, to the neuron to reset $V_{mem}$ and to implement the refractory period.

Figure 3b shows bursting behaviour. Here, the durations of the four excitatory spikes were set to 32 ns and the EPSP is strong enough to cause tonic bursting. $V_{mem}$ is only reset by the FPGA once five consecutive spikes have been received. Figure 3c shows spike frequency adaptation. Here, an excitatory spike is sent to the neuron every 1 ms for continuous stimulation. When an output spike is created, $V_{mem}$ is reset by sending RST to the neuron, which initially lasts 3 ns. Spike frequency adaptation is obtained by incrementing the duration of the RST pulse by 3 ns each time until it reaches a maximum duration of 32 ns. Figure 3d shows the effect of inhibitory spikes (at a finer time resolution than the previous results). In this case, the first and the third spikes received are excitatory, whereas the second and fourth spikes are inhibitory, all with equal weight (i.e., a duration of 24 ns). The synapse is excited first, then inhibited, excited again and finally inhibited. The spike threshold is never reached.

### IV. CONCLUSIONS

We have presented an aVLSI implementation of a conductance-based silicon neuron based on a first-order low-pass filter core. This neuron achieves an extremely high density, measuring only 26.5 μm$^2$ in a 130 nm technology. Besides the high density, the proposed neuron can be easily reconfigured via a digital controller on an FPGA for different firing patterns observed in biological neurons, allowing great flexibility in spiking behaviour, even after fabrication. Another major advantage of the proposed neuron over existing solutions [3] is that it uses only one fixed set of biases, which will makes it practical for being used in mixed-signal platforms for reconfigurable large-scale neural networks.

## V. ACKNOWLEDGMENT

This work has been supported by the Australian Research Council Grant DP140103001. This work was inspired by the Capo Caccia Cognitive Neuromorphic Engineering Workshop 2014 and Telluride Neuromorphic workshop 2014.

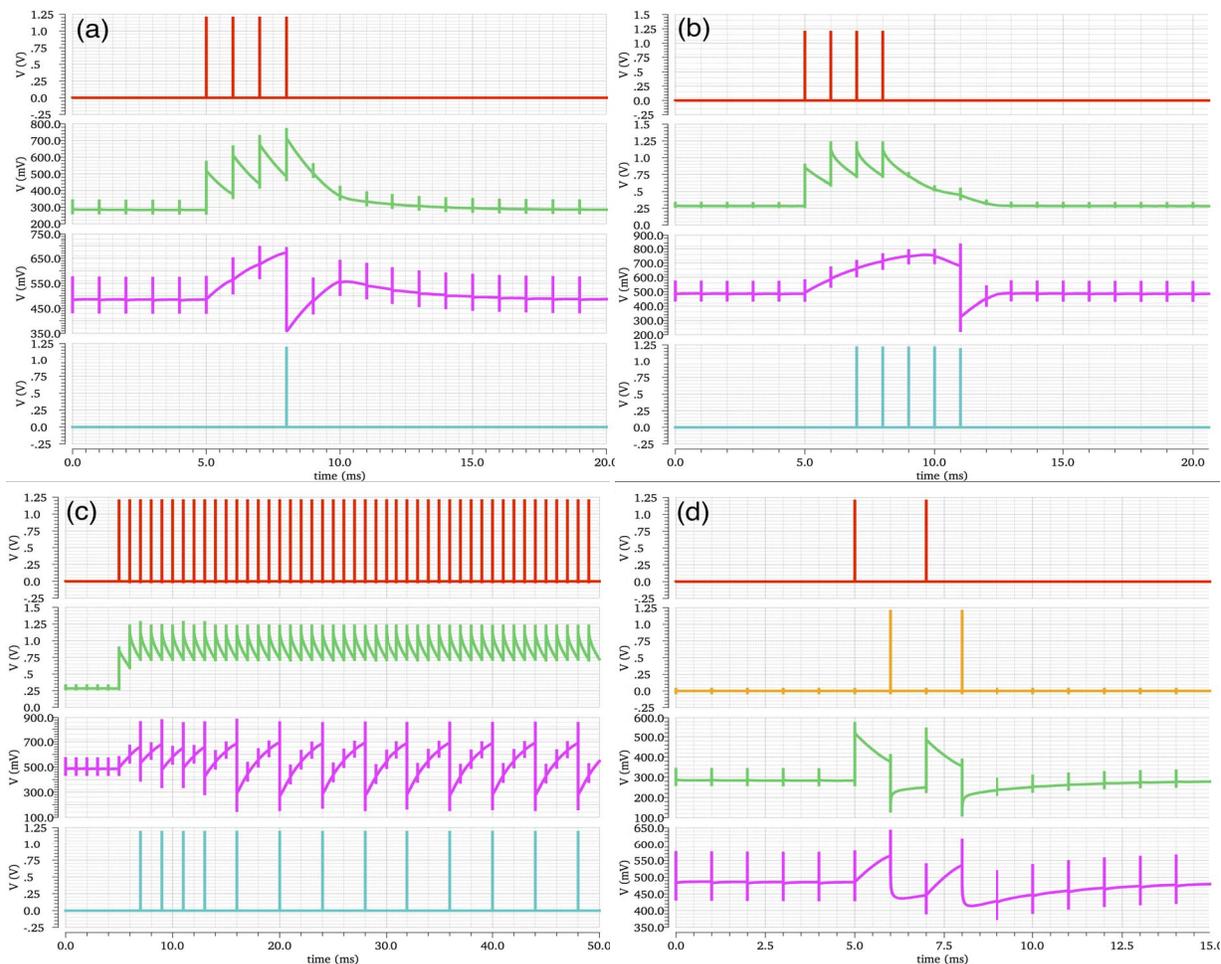

Figure 3. Simulation Results. (a) Tonic Firing; (b) Bursting; (c) Spike Frequency Adaptation; (d) Effect of inhibitory spikes. The red curve represents the excitatory spikes, the orange curve is the inhibitory spikes, the green curve represents $V_{syn}$, the pink curve represents $V_{mem}$ and the blue curve represents the spike generated by $I_{active}$.